\documentclass[fleqn,10pt]{wlscirep}
\usepackage[utf8]{inputenc}
\usepackage[T1]{fontenc}

\usepackage{lineno}
\usepackage{array}
\usepackage{longtable}
\usepackage{graphicx}
\usepackage{multirow}

\usepackage{fancyvrb}

\usepackage[nolist,nohyperlinks]{acronym}

\acrodef{us}[US]{ultrasound}
\acrodef{ai}[AI]{artificial intelligence}
\acrodef{sp}[SP]{standard plane}
\acrodef{hc}[HC]{head circumference}
\acrodef{ac}[AC]{abdominal circumference}
\acrodef{fl}[FL]{femur length}
\acrodef{bpd}[BPD]{bi-parietal diameter}
\acrodef{ofd}[OFD]{occipito-frontal diameter}
\acrodef{tad}[TAD]{transverse abdominal diameter}
\acrodef{apad}[APAD]{anterior-posterior abdominal diameter}
\acrodef{fp}[FP]{Fetal Plane}
\acrodef{uclh}[UCLH]{University College London Hospital}
\acrodef{dod}[DOD]{Dynamic Orientation Determination}
\acrodef{cnn}[CNN]{convolutional neural network}
\acrodef{dl}[DL]{deep learning}
\acrodef{ga}[GA]{gestational age}
\acrodef{tv}[TV]{transventricular}
\acrodef{ta}[TA]{transabdominal}
\acrodef{ml}[ML]{machine learning}
\acrodef{nme}[NME]{Normalised Mean Error}
\acrodef{via}[VIA]{VGG Image Annotator}
\acrodef{2d}[2D]{two-dimensional}
\acrodef{ucl}[UCL]{University College London}
\acrodef{m-c}[M-C]{MULTI-CENTRE}

\title{A multi-centre, multi-device benchmark dataset for landmark-based comprehensive fetal biometry}

\author[1,*]{Chiara Di Vece}
\author[1]{Zhehua Mao}
\author[2,3]{Netanell Avisdris}
\author[1]{Brian Dromey}
\author[4]{Raffaele Napolitano}
\author[2,5]{Dafna Ben Bashat}
\author[1]{Francisco Vasconcelos}
\author[1]{Danail Stoyanov}
\author[3]{Leo Joskowicz}
\author[1]{Sophia Bano}
\affil[1]{Department of Computer Science and UCL Hawkes Institute, University College London, London WC1E 6BT, UK}
\affil[2]{Sagol Brain Institute, Tel Aviv Sourasky Medical Center, Tel Aviv, Israel}
\affil[3]{School of Computer Science and Engineering, The Hebrew University of Jerusalem, Jerusalem, Israel}
\affil[4]{UCLH NHS Foundation Trust and the Elizabeth Garrett Anderson Institute for Women's Health, UCL, London, UK}
\affil[5]{Sagol School of Neuroscience and Sackler Faculty of Medicine, Tel Aviv University, Tel Aviv, Israel}
\affil[*]{Corresponding author: Chiara Di Vece (chiara.divece.20@ucl.ac.uk)}


\begin{abstract}
Accurate fetal growth assessment from \ac{us} relies on precise biometry measured by manually identifying anatomical landmarks in standard planes. Manual landmarking is time-consuming, operator-dependent, and sensitive to variability across scanners and sites, limiting the reproducibility of automated approaches. There is a need for multi-source annotated datasets to develop artificial intelligence-assisted fetal growth assessment methods. 
To address this bottleneck, we present an open, multi-centre, multi-device benchmark dataset of fetal \ac{us} images with expert anatomical landmark annotations for clinically used fetal biometric measurements. These measurements include head bi-parietal and occipito-frontal diameters, abdominal transverse and antero-posterior diameters, and femoral length.
The dataset contains 4,513 de-identified \ac{us} images from 1,904 subjects acquired at three clinical sites using seven different \ac{us} devices. 
We provide standardised, subject-disjoint train/test splits, evaluation code, and baseline results to enable fair and reproducible comparison of methods. Using an automatic biometry model, we quantify domain shift and demonstrate that training and evaluation confined to a single centre substantially overestimate performance relative to multi-centre testing. 
To the best of our knowledge, this is the first publicly available multi-centre, multi-device, landmark-annotated dataset that covers all primary fetal biometry measures, providing a robust benchmark for domain adaptation and multi-centre generalisation in fetal biometry and enabling more reliable AI-assisted fetal growth assessment across centres. 
All data and annotations are available on the \href{https://doi.org/10.5522/04/30819911}{UCL Research Data Repository}. Training code and evaluation pipelines are available at \href{https://github.com/surgical-vision/Multicentre-Fetal-Biometry.git}{https://github.com/surgical-vision/Multicentre-Fetal-Biometry.git}. 
\end{abstract}

\begin{document}

\flushbottom
\maketitle

\thispagestyle{empty}

\section*{Introduction}
\label{sec:introduction}

Fetal growth assessment requires precise biometric measurements obtained by manual identification of anatomical landmarks on \acp{sp}. \ac{us}-based estimation of fetal biometry is widely used to monitor fetal growth relative to expected \ac{ga} and to diagnose prenatal abnormalities. Reliable measurement requires sonographers to accurately identify anatomical landmarks on specific \acp{sp}~\cite{burgos2020zdataset}, predefined cross-sections defined by orientation, slice position, and presence of specific anatomical structures to ensure standardised fetal assessment across scans and users. Common \acp{sp} include the \ac{tv} \ac{sp} to measure fetal \ac{hc}, the \ac{ta} \ac{sp} to measure fetal \ac{ac}, and the femoral plane to measure \ac{fl}. 
However, manual landmark annotation is time-consuming, operator-dependent, and sensitive to variability across scanners and sites, thereby limiting reproducibility. Obtaining manual measurements, especially for trainees, results in high inter- and intra-operator variability~\cite{dromey2020dimensionless,joskowicz2019observervar}. In fetal biometry, inter-operator measurement variability ranges between $\pm$4.9\% and $\pm$11.1\%, and intra-operator variability between $\pm$3\% and $\pm$6.6\%~\cite{sarris2012intra}. This variability contributes to uncertainty in biometry and hampers the evaluation of fetal growth in clinical practice.

While \ac{dl} offers promise for automating medical imaging, model performance ultimately depends on the quality and diversity of the training data. Various approaches for automated fetal biometry have been proposed, ranging from single-measurement methods~\cite{torres2022headus_review}, including \ac{bpd}~\cite{al2019bpdauto,zhang2017automatic_fhc}, \ac{hc}~\cite{van2018automated}, and \ac{fl}~\cite{khan2017bpdfl}, to comprehensive multi-measurement frameworks for fetal weight estimation~\cite{bano2021autofb,prieto2021automated,avisdris2022biometrynet, plotka2021fetalnet}. 
While segmentation-based methods~\cite{bano2021autofb,plotka2021fetalnet} are time-consuming to annotate, direct landmark detection~\cite{salomon2019isuog} aligns better with clinical workflow, requiring sonographers to mark two anatomical points per measurement, representing a faster and more intuitive task~\cite{prieto2021automated}.

Despite progress in automated biometry, various \ac{us} machines (General Electric, Philips, Hitachi) introduce significant source variability, which can cause models trained on data from one clinical site to perform poorly when deployed at another, a phenomenon known as domain shift. Developing robust models has been hindered by the lack of large-scale datasets capturing variability across patients, operators, and \ac{us} devices.

The development of automatic methods for computer-aided diagnosis, especially \ac{dl}-based approaches, relies on the availability of relevant annotated datasets. Public datasets such as BRATS~\cite{menze2014multimodal} for brain lesion segmentation and FeTA~\cite{payette2021automatic} for fetal brain segmentation have driven innovation in medical imaging. A recent review of automatic \ac{us} analysis~\cite{Fiorentino2022fetalreview} noted that there is no comprehensive dataset for fetal biometry that includes all measurements needed for fetal weight estimation. Table~\ref{tab:datasets} summarises existing datasets for \ac{us}-based fetal anatomy and biometry, highlighting that no prior work provides a comprehensive, multi-centre, open-access dataset with landmark annotations for all biometric planes (head, abdomen, femur). 

\begin{table}[t]
\centering
\caption{Summary of papers describing the fetal \ac{us} biometry datasets. Note that papers that reuse annotated datasets from previous works are not described here. For all \acp{sp} we count both the number of images and fetuses $(\#I/\#F)$.}
\label{tab:datasets}
\resizebox{\linewidth}{!}{%
\begin{tabular}{|c|c|c|c|c|c|c|c|} 
\hline
\textbf{Paper} & \textbf{Public?} & \textbf{Head SPs} & \textbf{Abdomen SPs} & \textbf{Femur SPs} & \textbf{CRL SPs} & \textbf{US Devices} & \textbf{Methodology} \\ 
\hline
\cite{zhang2017automatic_fhc} & No & 41/41 (training 20/20; test 21/21) & N/A  & N/A & N/A  & \begin{tabular}[c]{@{}c@{}}Aplio 780, 790, 500 (Toshiba), \\Voluson 730 (GE)\end{tabular} & \begin{tabular}[c]{@{}c@{}}Texton-based segmentation \\ with Random Forest classification \end{tabular}  \\ 
\hline
\cite{khan2017bpdfl} & No & 273/– (training 106/-; test 167/-) & N/A & 321/– (training 124/-; test 197/-) & N/A  & \begin{tabular}[c]{@{}c@{}}Voluson E6, E8, 730, vivid q (GE)\\Acuson Antares Premium Edition (Siemens)\\HI VISION Preirus EUB-8500 (Hitachi)\end{tabular} & \begin{tabular}[c]{@{}c@{}}Hybrid Geometric Model-Based Computer Vision \\ with Deformable Model Refinement \end{tabular}  \\ 
\hline
\cite{van2018automated} & Yes & \begin{tabular}[c]{@{}c@{}}1334/551 (training 999/806; test 335/335)\end{tabular} & N/A & N/A & N/A & \begin{tabular}[c]{@{}c@{}}Voluson E8, 730 (GE)\end{tabular} & \begin{tabular}[c]{@{}c@{}}Random Forest pixel classification \\ with geometric model-based segmentation \end{tabular} \\ 
\hline
\cite{sinclair2018human} & No & \begin{tabular}[c]{@{}c@{}}2724/– (training 1948/–;\\ validation 216/–; test 539/–)\end{tabular} & N/A & N/A & N/A & Voluson E8 (GE) & \begin{tabular}[c]{@{}c@{}}DL (FCN-based semantic segmentation\\ with ellipse fitting)\end{tabular} \\ 
\hline
\cite{zhu2021automatic} & No & N/A & N/A & \begin{tabular}[c]{@{}c@{}}436 (augmented to 2,610)/–\\ (training 2300/–; test 310/–)\end{tabular} & N/A & Not specified & \begin{tabular}[c]{@{}c@{}}Random Forest regression (endpoint localization) \\ + DL (SegNet segmentation)\end{tabular} \\ 
\hline
\cite{oghli2021automatic} & Partial & \begin{tabular}[c]{@{}c@{}}1334/551 (HC18)\\ (Training: 999/806; test 335/335)\end{tabular} & 158/– & 315/– & N/A & \begin{tabular}[c]{@{}c@{}}Voluson E8, 730 (GE) (HC18)\\ Voluson E10 (GE) (local)\end{tabular} & \begin{tabular}[c]{@{}c@{}}DL (Attention MFP-Unet for segmentation) \\ with preprocessing\end{tabular} \\ 
\hline
\cite{prieto2021automated} & No & \begin{tabular}[c]{@{}c@{}}7274/– (training: 5819/-;\\validation 1646/–;\\ test: 1455/-)\end{tabular} & \begin{tabular}[c]{@{}c@{}}6717/– (training: 5374/-;\\ validation 2622/- ;\\ test: 1343/-)\end{tabular} & \begin{tabular}[c]{@{}c@{}}7216/– (training: 5773/-;\\validation 2466/–;\\ test: 1443/-)\end{tabular} & \begin{tabular}[c]{@{}c@{}}3152/– (training: 2522/-;\\validation 499/–;\\ test: 630/-)\end{tabular} & Not specified & \begin{tabular}[c]{@{}c@{}}DL (caliper removal, classification,\\segmentation, measurement)\end{tabular} \\
\hline
\cite{plotka2021fetalnet} & No & \begin{tabular}[c]{@{}c@{}}32215/700 (training 19329/420; \\ validation 6443/140; \\ test 6443/140)\end{tabular} & \begin{tabular}[c]{@{}c@{}}26403/700 (training 15842/420; \\ validation 5281/140; \\ test 5280/140)\end{tabular} & \begin{tabular}[c]{@{}c@{}}3706/700 (training 2224/420; \\ validation 741/140; \\ test 741/140)\end{tabular} & N/A & \begin{tabular}[c]{@{}c@{}}Voluson E8, E10, \\S6, S8, P8 (GE)\end{tabular} & \begin{tabular}[c]{@{}c@{}}Multi-task DL (spatio-temporal video analysis \\ with attention gates and stacked modules)\end{tabular} \\ 
\hline
\cite{bano2021autofb} & No & 135/42 (4-fold cross-validation) & 103/42 (4-fold cross-validation)  & 108/42 (4-fold cross-validation) & N/A  & Voluson (GE) & \begin{tabular}[c]{@{}c@{}}DL (multi-class semantic\\segmentation + ellipse/bbox fitting)\end{tabular} \\ 
\hline
\cite{avisdris2022biometrynet} & Yes & \begin{tabular}[c]{@{}c@{}}999/806 (HC18)\\ 1638/909 (FP)\end{tabular} & N/A & 761/630 (FP) & N/A & \begin{tabular}[c]{@{}c@{}}Voluson E8, 730 (GE) (HC18)\\ Voluson E6, S8, S10 (GE), Aloka (FP)\end{tabular} & \begin{tabular}[c]{@{}c@{}}DL (landmark regression with \\Dynamic Orientation Determination)\end{tabular}\\
\hline
\end{tabular}
}
\end{table}

Recent advances in automated fetal biometry underscore the critical importance of diverse, annotated datasets. For instance, AutoFB~\cite{bano2021autofb} demonstrated the feasibility of automated biometry estimation; however, like most existing approaches, it was trained on limited, single-site data, underscoring the need for benchmarking across diverse acquisition conditions.

Similarly, Kim et al.~\cite{kim2019automatic} and Oghli et al.~\cite{oghli2020automatic} developed methods for automatic evaluation of fetal head biometry using \ac{ml} and deep neural networks, respectively, demonstrating the feasibility of automating key measurements such as \ac{hc} and \ac{bpd}. More recently, Slimani et al.~\cite{slimani2023fetal} and Lee et al.~\cite{lee2023machine} explored end-to-end automation of fetal biometry and \ac{ga} estimation using \ac{dl}, emphasising the need for diverse and multi-source datasets to address variability in \ac{us} devices and operator expertise. Other approaches have validated automated biometry at scale: Venturini et al.~\cite{venturini2025whole} achieved human-level performance through Bayesian frame-aggregation on 1,457 scans, while multi-country studies~\cite{benson2025fetal} demonstrated robust \ac{ga} estimation (1.7--2.8 days error) across 78,000 pregnancies in Australia, India, and the UK. Complementary work~\cite{goetz2025development} developed real-time systems for automatic HC, AC, and FL measurement with temporal validation across diverse acquisition conditions. These breakthroughs underscore the critical need for diverse, publicly available benchmark datasets to enable fair comparison and out-of-distribution validation.

To address this bottleneck, this paper presents the first open-access, multi-centre, multi-device benchmark dataset of fetal \ac{us} images with expert anatomical landmark annotations for all clinically used biometric measurements: \ac{bpd} and \ac{ofd} acquired on the fetal head plane, \ac{tad} and \ac{apad} acquired on the fetal abdominal plane, and \ac{fl} acquired on the fetal femur plane. The dataset combines three sources: the \ac{fp} dataset~\cite{burgos2020zdataset}, the HC18 head dataset~\cite{van2018automated}, and an expanded cohort from \ac{uclh}~\cite{bano2021autofb} (UCL) containing a total of 4,513 images from 1,904 subjects acquired at three clinical sites with seven different \ac{us} devices (Table~\ref{tab:datasets_summary}). We provide standardised landmark annotations, subject-disjoint train/validation/test splits, and technical validation demonstrating that multi-centre training substantially improves generalisation to unseen acquisition conditions. We demonstrate the technical usability of these annotations for automatic fetal biometry in a \ac{dl} framework~\cite{avisdris2022biometrynet} suitable for landmark-based annotation. This study evaluates landmark placement variability exclusively within pre-selected, clinically appropriate \acp{sp} as all images in our dataset were acquired by experienced sonographers who had already identified and captured the correct anatomical plane in accordance with ISUOG guidelines. 

\section*{Results}
\label{sec:results}

\subsection*{Quantifying anatomical variability across sites}

\begin{figure}[t]
\centering
\includegraphics[width=0.9\linewidth]{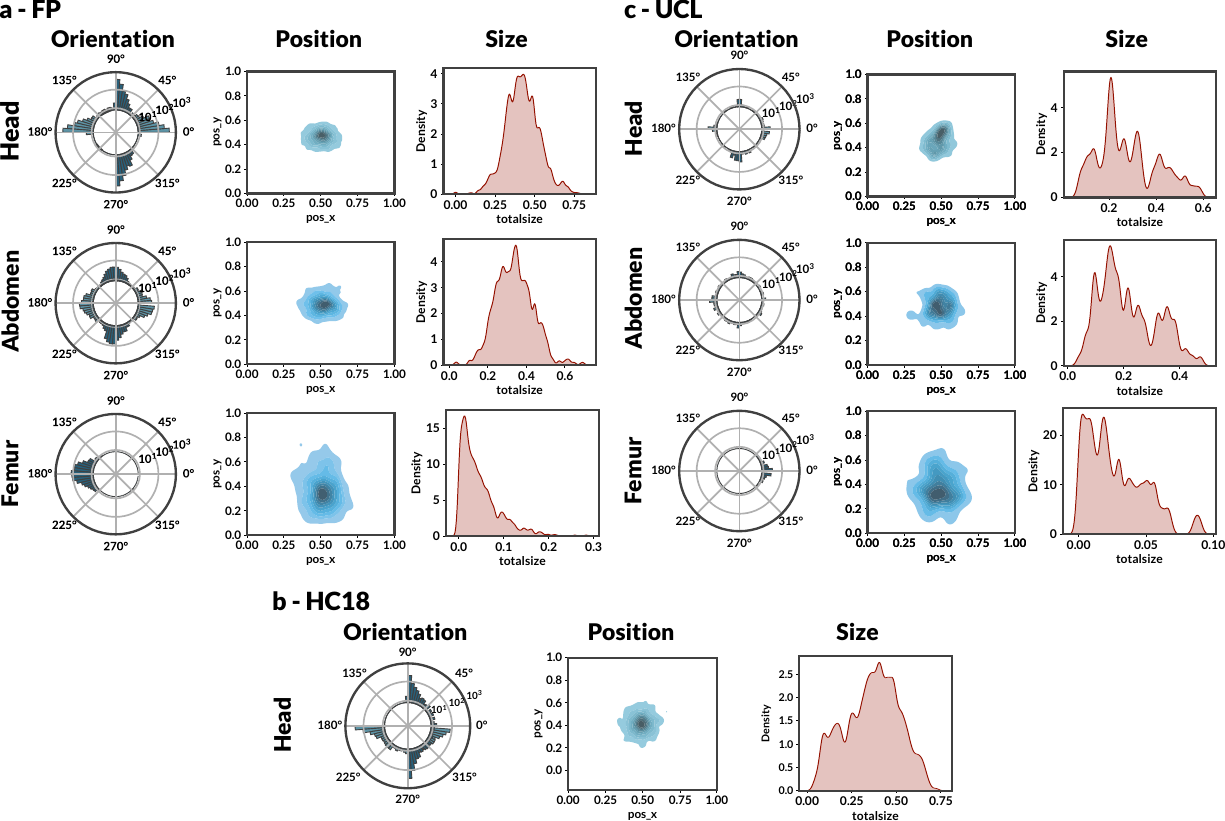}
\caption{Variability of anatomical structures across a) FP, b) HC18, and c) UCL datasets. Each row represents one anatomical region. \textit{Orientation}: polar histogram with log density scale of measurement angle relative to horizontal axis ([0$^{\circ}$,360$^{\circ}$]); \textit{Position}: 2D kernel density estimation of centre-point location (pos\_x, pos\_y $\in$ [0,1]); \textit{Size}: 1D kernel density estimation of structure size normalised by image area (unitless). Substantial heterogeneity across datasets reflects realistic clinical variability and demonstrates the domain-shift problem.}
\label{fig:datadist}
\end{figure}

We quantified anatomical variability across sites by analysing structure position (centre-point distribution), size (area relative to the image), and orientation (angle from horizontal). The analysis is shown in Figure~\ref{fig:datadist}.   

Figure~\ref{fig:datadist} reveals substantial intra- and inter-dataset variability in structure position, size, and orientation. Notably, structures are inconsistently centred due to differences in operator-dependent framing and probe handling. Size variability indicates heterogeneous magnification levels, with HC18 exhibiting the tightest range because all images were collected on a single scanner under a controlled protocol. In contrast, \ac{fp} and UCL include multiple devices and operators. Orientation differences are most evident in the femur plane, where the long bone may appear at any angle depending on fetal pose. These variations reflect the realistic heterogeneity encountered in clinical practice and demonstrate the domain-shift problem: models trained on single-site data fail to generalise to the broader distribution of poses, scales, and orientations in multi-centre data. This variability directly explains the performance degradation observed in cross-site evaluation.

\subsection*{Technical validation on benchmark datasets}

To evaluate the usability of the proposed dataset and annotation method, we use BiometryNet~\cite{avisdris2022biometrynet}, an end-to-end landmark regression framework for fetal biometry estimation, following the original evaluation protocol. It uses a \ac{dod} method to enforce measurement-specific orientation consistency during training on a modified HRNet architecture. This provides quantitative evidence of the dataset's value for developing automated biometry methods.

\subsubsection*{Cross-dataset performance}

\begin{figure}[t]
    \centering
    \includegraphics[width=0.9\linewidth]{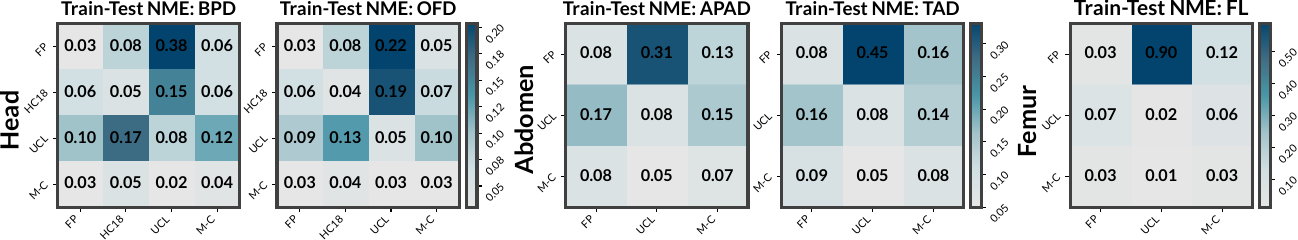}
    \caption{Cross-dataset generalisation heatmaps for fetal biometry. Train$\rightarrow$Test NME (lower is better) for each biometric measurement. Rows denote the training dataset and columns the test dataset. For abdomen and femur, HC18 is omitted where results are unavailable in the cross-dataset evaluation table. Cell values report mean NME. Colour scales are shared within each anatomy group (head: BPD/OFD; abdomen: APAD/TAD; femur: FL).}
    \label{fig:error-matrix}
\end{figure}

We performed a comprehensive evaluation across all three datasets and the combined multi-centre dataset (\ac{m-c}). For each measurement, localisation error is quantified using the \ac{nme} in image space: $\mathrm{NME}^{(i)} = (\min\bigl(d_{\mathrm{std}}^{(i)}, d_{\mathrm{swap}}^{(i)}\bigr)) / (2\,\lVert \mathbf{p}_1^{(i)} - \mathbf{p}_2^{(i)} \rVert_2)$, where $d_{\mathrm{std}}^{(i)}$ and $d_{\mathrm{swap}}^{(i)}$ are the sums of endpoint errors under standard and swapped correspondences, respectively, making the metric invariant to endpoint ordering. Table~\ref{tab:cross_validation} reports the mean and standard deviation of $\mathrm{NME}^{(i)}$ across the test set, demonstrating expected domain shift across imaging devices and clinical sites.
\begin{table}
\centering
\caption{Cross-data evaluation results showing NME$\pm$STD for all train--test combinations across four datasets (FP, HC18, UCL, M-C) and three anatomies. NME is unitless (measurement error normalised by inter-landmark distance). Multi-centre models are trained on combined multi-centre data. Within each \emph{training dataset block} and for each biometric measurement, \textbf{bold} indicates the best (lowest) NME across the test datasets in that block, and \underline{underline} indicates the second-best.}
\label{tab:cross_validation}
\begin{tabular}{|c|c|c|c|c|c|c|}
\hline
\multirow{3}{*}{\textbf{Train}} & \multirow{3}{*}{\textbf{Test}} & \multicolumn{5}{c|}{\textbf{NME $\pm$ STD}} \\
\cline{3-7}
 &  & \multicolumn{2}{c|}{Head} & \multicolumn{2}{c|}{Abdomen} & Femur \\
\cline{3-7}
 &  & BPD & OFD & APAD & TAD & FL \\
\hline
\multirow{4}{*}{FP} & FP
& \textbf{0.03$\pm$0.06} & \textbf{0.03$\pm$0.05} & \textbf{0.08$\pm$0.06} & \textbf{0.08$\pm$0.06} & \textbf{0.03$\pm$0.11} \\
\cline{2-7}
& HC18
& 0.08$\pm$0.12 & 0.08$\pm$0.13 &  &  &  \\
\cline{2-7}
& UCL
& 0.38$\pm$0.26 & 0.22$\pm$0.22 & 0.31$\pm$0.23 & 0.45$\pm$0.28 & 0.90$\pm$0.54 \\
\cline{2-7}
& \ac{m-c}
& \underline{0.06$\pm$0.14} & \underline{0.05$\pm$0.10} & \underline{0.13$\pm$0.15} & \underline{0.16$\pm$0.21} & \underline{0.12$\pm$0.34} \\
\hline
\multirow{4}{*}{HC18} & FP
& \underline{0.06$\pm$0.07} & \underline{0.06$\pm$0.07} &  &  &  \\
\cline{2-7}
& HC18
& \textbf{0.05$\pm$0.09} & \textbf{0.04$\pm$0.08} &  &  &  \\
\cline{2-7}
& UCL
& 0.15$\pm$0.16 & 0.19$\pm$0.23 &  &  &  \\
\cline{2-7}
& \ac{m-c}
& 0.06$\pm$0.11 & 0.07$\pm$0.11 &  &  &  \\
\hline
\multirow{4}{*}{UCL} & FP
& \underline{0.10$\pm$0.11} & \underline{0.09$\pm$0.09} & 0.17$\pm$0.13 & 0.16$\pm$0.12 & 0.07$\pm$0.18 \\
\cline{2-7}
& HC18
& 0.17$\pm$0.25 & 0.13$\pm$0.16 &  &  &  \\
\cline{2-7}
& UCL
& \textbf{0.08$\pm$0.18} & \textbf{0.05$\pm$0.11} & \textbf{0.08$\pm$0.14} & \textbf{0.08$\pm$0.14} & \textbf{0.02$\pm$0.03} \\
\cline{2-7}
& \ac{m-c}
& 0.12$\pm$0.17 & 0.10$\pm$0.12 & \underline{0.15$\pm$0.14} & \underline{0.14$\pm$0.13} & \underline{0.06$\pm$0.17} \\
\hline
\multirow{4}{*}{\ac{m-c}} & FP
& \underline{0.03$\pm$0.05} & \textbf{0.03$\pm$0.04} & 0.08$\pm$0.06 & 0.09$\pm$0.07 & 0.03$\pm$0.10 \\
\cline{2-7}
& HC18
& 0.05$\pm$0.08 & 0.04$\pm$0.07 &  &  &  \\
\cline{2-7}
& UCL
& \textbf{0.02$\pm$0.02} & 0.03$\pm$0.11 & \textbf{0.05$\pm$0.12} & \textbf{0.05$\pm$0.12} & \textbf{0.01$\pm$0.01} \\
\cline{2-7}
& \ac{m-c}
& 0.04$\pm$0.07 & \underline{0.03$\pm$0.06} & \underline{0.07$\pm$0.08} & \underline{0.08$\pm$0.08} & \underline{0.03$\pm$0.09} \\
\hline
\end{tabular}
\end{table}
Figure~\ref{fig:error-matrix} visualises the same train--test \ac{nme} values as a set of heatmaps for each anatomy. This representation highlights the relative robustness of \ac{m-c} training across test datasets and the pronounced asymmetry of some cross-domain pairs, particularly for \ac{fl}.
Within-domain performance (training and testing on the same dataset) achieved low errors across all anatomies, with \ac{nme} values typically below 0.1 for head and abdomen (\ac{bpd}, \ac{ofd}, \ac{apad}, \ac{tad}) and below 0.05 for \ac{fl} (Table~\ref{tab:cross_validation}). Cross-domain performance (training on one dataset, testing on another) showed a consistent increase in \ac{nme}, particularly when source and target differed in device manufacturer or acquisition protocol. Averaged across all train--test pairs, head and abdomen \ac{nme} roughly doubled relative to within-domain performance, while femur \ac{nme} reached as high as 0.90 in the FP$\rightarrow$UCL setting, indicating substantial domain shift.

For head biometry (\ac{bpd} and \ac{ofd}), the \ac{m-c} model achieved the best overall cross-site generalisation. When trained on \ac{m-c} and tested on UCL, it achieved \ac{nme}~=~0.02$\pm$0.02 for \ac{bpd} and 0.03$\pm$0.11 for \ac{ofd}, outperforming even the UCL-trained model evaluated on its own test set (0.08$\pm$0.18 and 0.05$\pm$0.11, respectively). This demonstrates that incorporating diverse training data improves generalisation, even to datasets included in the training pool. After correcting preprocessing to ensure that all landmarks fall within the heatmap, HC18-trained models exhibit competitive within-domain performance and moderate cross-domain degradation, rather than the catastrophic failures observed before the preprocessing fix. For abdomen biometry (\ac{apad} and \ac{tad}), UCL- and \ac{m-c}-trained models achieved the lowest within-domain errors (UCL$\rightarrow$UCL: 0.08$\pm$0.14 for both \ac{apad} and \ac{tad}; \ac{m-c}$\rightarrow$\ac{m-c}: 0.07$\pm$0.08 and 0.08$\pm$0.08). Cross-domain transfer between FP and UCL showed a clear but more modest degradation than for the femur, with \ac{nme} increasing to 0.16--0.31 depending on the direction of transfer. For femur length, both FP and UCL models achieved excellent within-domain performance (\ac{nme}~$\approx$~0.03), but cross-domain transfer remained challenging, especially from FP to UCL (FP$\rightarrow$UCL: 0.90$\pm$0.54). The \ac{m-c}-trained model achieved the lowest femur errors on the \ac{m-c} test set (0.03$\pm$0.09), but cross-site generalisation for \ac{fl} remains limited. 

\begin{figure}[!t]
\centering
\includegraphics[width=0.95\linewidth]{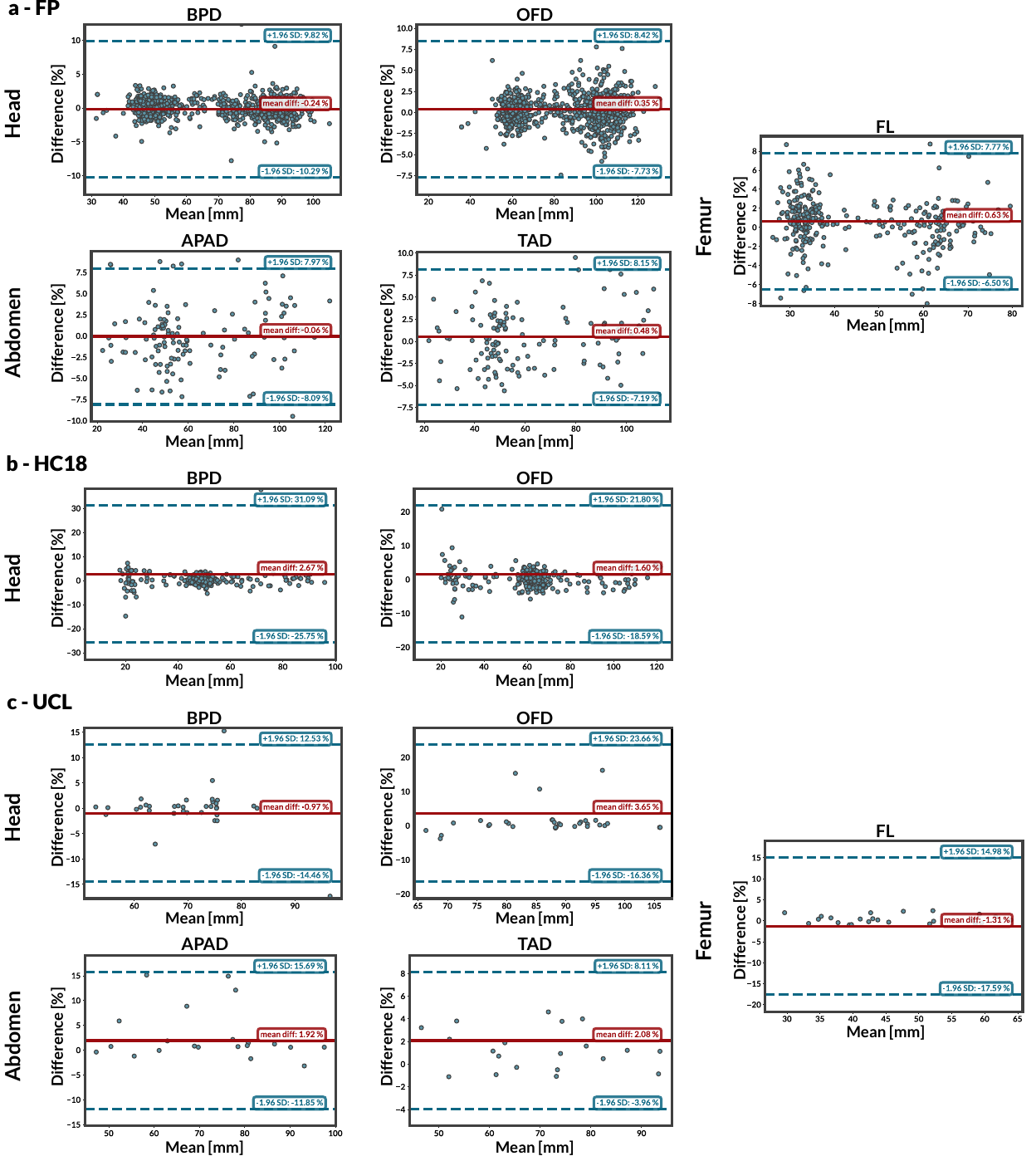}
\caption{Normalised Bland–Altman agreement plots between BiometryNet predictions and ground-truth fetal biometry measurements for a) FP, b) HC18, and c) UCL datasets. Horizontal axis: mean measurement (mm). Vertical axis: percent difference relative to the ground-truth measurement. Solid lines indicate the mean bias and dashed lines indicate the 95\% limits of agreement (bias $\pm 1.96$ SD).}
\label{fig:results}
\end{figure}

\begin{figure}[t]
    \centering
    \includegraphics[width=0.9\linewidth]{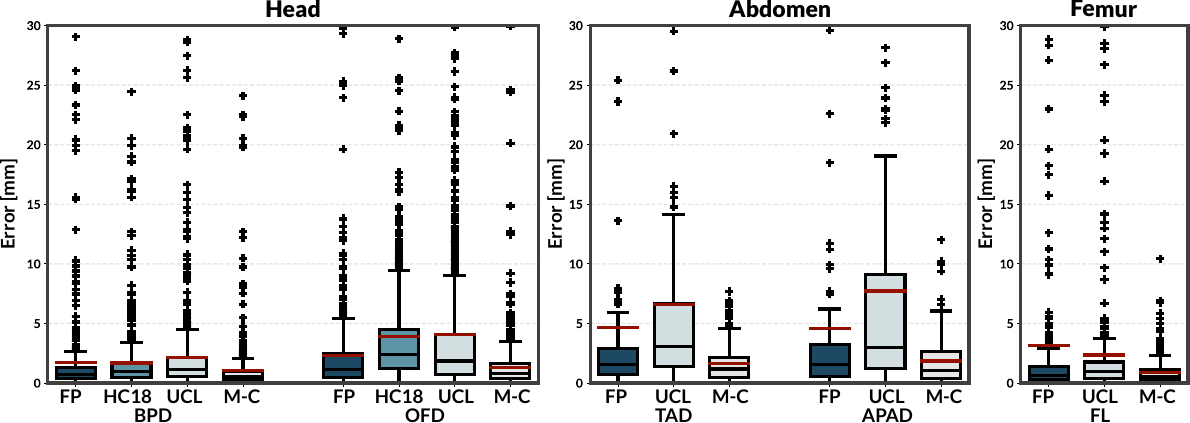}
    \caption{Absolute biometry error (mm) on the \ac{m-c} test set for models trained on FP, HC18, UCL, and \ac{m-c} datasets. Boxplots are shown for Head (BPD, OFD), Abdomen (TAD, APAD), and Femur (FL). The y-axis is truncated at 30\,mm to improve visualisation of the central distribution.}
    \label{fig:boxplots}
\end{figure}

Figure~\ref{fig:results} shows normalised Bland--Altman plots for each dataset, illustrating the agreement between predicted and ground-truth measurements as a \emph{percent difference} relative to the ground-truth. Expressing errors as percentages mitigates heteroscedasticity with increasing measurement size and enables comparison across anatomies. Solid lines indicate the mean bias and dashed lines indicate the 95\% limits of agreement (bias~$\pm$~1.96~SD). Distances are converted from pixels to millimetres before computing mean measurements and percentage differences.
Figure~\ref{fig:boxplots} shows absolute biometry error (mm) on the \ac{m-c} test set for models trained on FP, HC18, UCL, and \ac{m-c} for head, abdomen, and femur (\ac{fl}); to focus on the central error distribution, the y-axis is truncated at 30\,mm; a small number of larger outliers fall above this range. The \ac{m-c}-trained model achieves competitive medians across all anatomies with consistently narrow interquartile ranges, indicating robust and stable performance. In contrast, HC18-trained models exhibit wider error distributions on the \ac{m-c} test set, reflecting residual domain shift despite the corrected preprocessing. For head, the median errors of the \ac{m-c} model ($\approx$ 0.2\,mm for \ac{bpd} and \ac{ofd}) correspond to clinically acceptable \ac{ga} variation in late pregnancy.
Overall, these results demonstrate that models trained on multi-centre data generalise more robustly across clinical sites, establishing baseline performance benchmarks for future method development and highlighting the importance of multi-centre training for clinically deployable fetal biometry systems.

\subsection*{Data description}
The dataset comprises three distinct subsets acquired from three clinical sites (Table~\ref{tab:datasets_summary}). HC18 spans all trimesters with a clinical screening bias towards the second trimester. FP and UCL datasets are predominantly second/third trimester (14--40 weeks), capturing the \ac{ga} range used for fetal weight estimation. This diversity ensures the benchmark covers the full clinical spectrum of fetal biometry assessment. When combined for multi-centre training, we refer to the unified dataset as \ac{m-c} in cross-data evaluation experiments.

\subsubsection*{File organisation}
The dataset is organised into three main subsets, each contained in a separate directory with a standardised structure. Each dataset directory contains a comprehensive README.md file that provides detailed information on dataset characteristics, acquisition protocols, device-specific details, imaging parameters, annotation conventions, measurement definitions, data preprocessing steps, and quality control procedures. The primary repository README.md provides an overview of the entire dataset collection and guidance on selecting datasets based on research objectives. Each dataset subset follows a standardised organisation:

\begin{Verbatim}[fontsize=\footnotesize]
[Dataset]/
|-- annotations/
|   |-- [Anatomy].csv          # Complete annotations
|   |-- [Anatomy]_Train.csv    # Training split
|   `-- [Anatomy]_Test.csv     # Test split
`-- data/
    |-- Head/                  # Head ultrasound images
    |-- Abdomen/               # Abdomen ultrasound images (if applicable)
    `-- Femur/                 # Femur ultrasound images (if applicable)
\end{Verbatim}

where \texttt{[Dataset]} is one of \{FP, HC18, UCL\} and \texttt{[Anatomy]} is one of \{Head, Abdomen, Femur\}.

\subsubsection*{Annotation format}

Annotations are provided as CSV files, each containing landmark coordinates for a single measurement. All CSV files include common fields \textit{index} (sequential index in the dataset), \textit{image\_name} (filename of the corresponding \ac{us} image), \textit{scale} (scaling factor applied during preprocessing), \textit{center\_w, center\_h} (centre coordinates of the region of interest), \textit{px\_to\_mm\_rate} (pixel-to-millimetre conversion rate), \textit{mm\_dist} (distance marker value visible in the image, typically 5mm or 10mm), \textit{Algo} (device or algorithm identifier), \textit{SubjectID} (de-identified subject identifier), \textit{Split} (data split indicator, train or test).

\textbf{Head Measurements (BPD and OFD)}: \texttt{bpd\_1\_x, bpd\_1\_y, bpd\_2\_x, bpd\_2\_y} (landmarks for bi-parietal diameter) and \texttt{ofd\_1\_x, ofd\_1\_y, ofd\_2\_x, ofd\_2\_y} (landmarks for occipito-frontal diameter). 

\textbf{Abdomen Measurements (TAD and APAD)}:\texttt{tad\_1\_x, tad\_1\_y, tad\_2\_x, tad\_2\_y} (landmarks for transverse abdominal diameter) and \texttt{apad\_1\_x, apad\_1\_y, apad\_2\_x, apad\_2\_y} (landmarks for anterior-posterior abdominal diameter).

\textbf{Femur Measurements (FL)}: \texttt{fl\_1\_x, fl\_1\_y, fl\_2\_x, fl\_2\_y} (landmarks for femur length, calculated between proximal and distal ends).

\subsubsection*{Data splits}

All subsets provide standardised, subject-disjoint train/test splits to ensure unbiased evaluation. Images from the same subject appear in only one split (either train or test). The split assignments are provided in both separate files (\texttt{[Anatomy]\_Train.csv} and \texttt{[Anatomy]\_Test.csv}) and indicated in the \texttt{Split} column of the complete annotation files.

\section*{Discussion}
\label{sec:discussion}

We present the first comprehensive, open-access, multi-centre dataset for landmark-based fetal biometry automation. By combining three sources (FP, HC18, UCL) across 7 \ac{us} devices and 3 clinical sites, we provide 4,513 annotated images that capture realistic clinical variability. Unlike single-anatomy (HC18) or single-site (AutoFB) datasets, our dataset enables systematic quantification of domain shift, a critical barrier to clinical deployment of \ac{ai}-based biometry systems. Our dataset and experiments focus on landmark detection within pre-validated \acp{sp}. The variability introduced by \ac{sp} selection was not quantified in this work. While this approach isolates variability in landmark placement for rigorous benchmarking, a fully automated pipeline would need to address either plane verification or robustness to plane variations. Future work should extend benchmarking to include raw ultrasound sweeps or near-boundary planes to quantify \ac{sp} selection variability and provide a pathway toward fully autonomous fetal biometry.

Table~\ref{tab:cross_validation} quantifies the performance gap between within-dataset and cross-dataset evaluation. Within-domain evaluation achieved low \ac{nme} values (typically below 0.1 for head and abdomen measurements), whereas cross-domain evaluation showed a marked increase in error, with some train--test pairs more than doubling the \ac{nme} and femur cross-domain \ac{nme} reaching values close to 1.0 in the most challenging settings (e.g., FP$\rightarrow$UCL). This domain shift arises from differences in imaging devices (GE Voluson E8 vs.\ Aloka), operator framing conventions (Figure~\ref{fig:datadist}), and fetal presentation variability. The domain shift is the primary barrier to clinical deployment, as models trained on single-site data are unlikely to maintain accuracy when deployed in different clinical settings.

Cross-dataset training required standardising landmark ordering conventions to ensure consistent supervision signals. Different datasets employed opposite endpoint conventions for certain measurements (e.g., FP annotated \ac{fl} with pt$_1$ on the right and pt$_2$ on the left, whereas UCL used the opposite convention). We leveraged BiometryNet's \ac{dod} mechanism, which learns a measurement-specific orientation vector during training. We extended the framework to save this vector in model checkpoints and apply it at inference time, automatically correcting predicted landmarks to match the training dataset's endpoint ordering convention. This approach avoids manual point-swapping and ensures fair cross-dataset comparison; without this correction, cross-dataset errors were artificially inflated because many predictions were swapped endpoint pairs that represented correct localisation but mismatched ordering conventions.

Measurement accuracy varies with the \ac{ga}. HC18 analysis~\cite{van2018automated} showed \ac{ga} estimation errors increased from the first trimester (0.6$\pm$4.3 days, ~0.7\% deviation) to the third (2.5$\pm$12.4 days, ~1.1\% deviation), reflecting increased biological variability and varying pixel magnification across \ac{us} devices. FP and UCL datasets capture this real-world \ac{ga} variability across multiple devices and operators, though trimester-specific error stratification was not quantified in the original studies.

The \ac{m-c} model achieved the best performance when tested on UCL data (\ac{bpd}: 0.02$\pm$0.02; \ac{ofd}: 0.03$\pm$0.11), outperforming even the UCL-trained model tested on its own data (\ac{bpd}: 0.08$\pm$0.18; \ac{ofd}: 0.05$\pm$0.11). This demonstrates that incorporating diverse training data improves generalisation, even to datasets included in the training set. However, the \ac{m-c} model showed limited cross-site generalisation for femur measurements on the \ac{m-c} test set (0.03$\pm$0.09), and inspection of the predictions revealed that many high-\ac{nme} failures correspond to the model selecting the wrong femoral instance when more than one long bone is visible in the image, rather than misplacing endpoints along the correct bone. Since \ac{nme} is normalised by the ground-truth inter-landmark distance (approximately the femur length), an error on the order of the distance between the two femurs produces \ac{nme} values close to 1.0. Thus, the heavy tail in cross-domain \ac{fl} errors reflects instance ambiguity rather than small-scale localisation noise, indicating that robust femur biometry may require anatomy-specific strategies (e.g., region-of-interest priors or multiple-instance learning).

Before correcting the preprocessing, the HC18 dataset showed markedly inferior cross-domain generalisation due to landmark-space mismatch. After applying image-centric preprocessing parameters, HC18-trained models show competitive within-domain performance and moderate cross-domain degradation (Table~\ref{tab:cross_validation}: HC18$\rightarrow$FP = 0.06$\pm$0.07, HC18$\rightarrow$UCL = 0.15$\pm$0.16). Residual differences in annotation methodology remain due to differing annotation procedures, underscoring the importance of harmonising annotation protocols in multi-centre dataset development.

These findings have important implications for clinical deployment. The marked performance degradation from within-domain to cross-domain settings indicates that models trained on single-site data are unlikely to maintain accuracy when deployed in different clinical settings with varying \ac{us} devices and operator practices. The improved cross-site generalisation of multi-centre models (e.g., \ac{m-c}$\rightarrow$UCL, achieving substantially lower \ac{nme} values than FP$\rightarrow$UCL for head biometry) suggests that robust clinical deployment will require training on diverse, multi-source datasets that capture real-world variability. For sonographers, successful automation could reduce measurement time and inter-operator variability (currently 4.9--11.1\% across observers)~\cite{sarris2012intra}, but only if models are validated across the full spectrum of clinical acquisition conditions encountered in practice.

By releasing this dataset publicly, we enable the research community to develop and benchmark automated fetal biometry methods under realistic clinical variability, addressing reproducibility challenges that have limited progress in this field. Our benchmarking infrastructure with standardised splits, training and evaluation code provides a foundation for reproducible research. Future work should extend this infrastructure in three directions: i) temporal sequences and video data to leverage anatomical motion consistency, ii) pathological cases and anomalous presentations to ensure robustness across clinical scenarios, and iii) additional anatomical planes (e.g., fetal abdomen, thorax) to enable comprehensive fetal assessment. By enabling robust, generalisable benchmarking across diverse clinical settings, this work lays the foundation for future advances in automated fetal biometry and improved clinical screening workflows.

\section*{Methods}
\label{sec:methods}
Our approach combines three existing datasets with different acquisition protocols, devices, and annotation strategies. We standardise annotations to a landmark-based format with consistent ordering conventions, compute image-specific preprocessing parameters for the 256$\times$256 model input, and provide subject-disjoint train/test splits for reproducible benchmarking. This section describes each source dataset, our annotation pipeline, preprocessing methodology, and the scale-recovery method for converting pixel measurements to clinical units.

\subsection*{Source dataset}
\label{subsec:datasets}

We use two publicly available datasets, \ac{fp}~\cite{burgos2020zdataset} and HC18~\cite{van2018automated}, and UCL, an in-house dataset from \ac{uclh} (Table~\ref{tab:datasets_summary}). 
\begin{table}[!t]
\centering
\caption{Summary of the datasets included in this study. Each dataset contains standard 2D US planes of the fetal head, abdomen, and femur, annotated with anatomical landmarks for biometry estimation, except for HC18, which only contains the fetal head.}
\label{tab:datasets_summary}
\resizebox{\columnwidth}{!}{
\begin{tabular}{| p{1.25cm}  |p{3cm}  |p{2cm}  |p{1.25cm}  |p{2.5cm} |p{2.50cm}  |p{4.50cm} |}
\hline
\textbf{Dataset} & \textbf{Source/Institution} & \textbf{\#Subjects} & \textbf{\#Images} & \textbf{Anatomical Planes} & \textbf{Annotation Tool} & \textbf{Notes} \\
\hline
\textbf{FP}~\cite{burgos2020zdataset} & Vall d'Hebron and Hospital Sant Joan de Déu, Barcelona & 1,047 & 3,090 & Head (1,637), Abdomen (693), Femur (760) & VIA & Six \ac{us} devices (GE Voluson E6, S8, S10, Aloka). Landmark annotations by obstetricians. \\ \hline
\textbf{HC18}~\cite{van2018automated} & Radboud University Medical Center, Netherlands & 806 & 999 & Head only (999) & Ellipse fitting from segmentation mask & BPD and OFD derived via least-square ellipse fitting from head circumference masks. \\ \hline
\textbf{UCL} & University College London Hospital, UK & 51 & 424& Head (159), Abdomen (130), Femur (135) & VIA & Approved under IRAS ID 230125. Pseudo-anonymised; NHS Fetal Anomaly Screening Programme compliant. \\
\hline
\textbf{\ac{m-c}} & Multi-centre & \textbf{1,904} & \textbf{4,513} & \textbf{Head (2,795), Abdomen (823), Femur (895)} &  &  \\
\hline
\end{tabular}}
\end{table}
\ac{fp}~\cite{burgos2020zdataset} was originally designed for the \ac{us} \acp{sp} classification challenge. The \ac{us} images were acquired on six \ac{us} devices: three GE Voluson E6, one Voluson S8, one Voluson S10, and one Aloka at two clinical sites in Barcelona, Spain. Since not all images in \ac{fp} qualified as \acp{sp} for fetal biometry~\cite{salomon2008score}, we selected 1,637 (909 subjects) fetal head, 693 (586 subjects) fetal abdomen, and 760 (629 subjects) fetal femur \acp{sp}. Many subjects contributed images across multiple anatomies, resulting in a total of 3,090 images from 1,047 unique subjects. An obstetrician then manually annotated the landmarks on each image with the \ac{via} annotation tool~\cite{dutta2019vgg}. HC18~\cite{van2018automated} was designed for the fetal \ac{hc} challenge. The \ac{us} images were acquired with two \ac{us} devices, GE Voluson E8 and 730, at a single clinical site in the Netherlands. All HC18 training sets, 999 (806 subjects) fetal head \acp{sp}, were annotated with a \ac{hc} measurement. We extracted the \ac{bpd} and \ac{ofd} biometric measurements from an ellipse's major and minor axes by least-square fitting~\cite{fitzgibbon1996ellipsefit} onto the ground-truth mask.

The local research ethics committee reviewed and approved the UCL dataset collection process (IRAS ID 230125). Patients attending the hospital for \ac{us} examination were enrolled and pseudo-anonymised by the clinical research staff. Each patient gave written consent. 
The complete image library from each \ac{us} was transferred to the research database. 
All \ac{us} images saved by the operator were considered optimal for that scan and of diagnostic quality. The measurement callipers were applied by the \ac{us} operator, and in most cases, images with and without the measurement callipers were saved. A clinical research fellow extracted a subset of images relevant to fetal biometry from the database. A total of 424 images were included from 51 pregnancies. Each image in the set data was classified as \ac{ac}, \ac{hc}, or \ac{fl}. The \ac{via} annotation tool~\cite{dutta2019vgg} was used to annotate the head, abdomen, or femur within each image for the segmentation task.
The fully anonymised standard \ac{us} plane images obtained exhibit significant intra-class variability. For example, in some cases the femur is well aligned with the plane's centre, while in others it appears as a small, distant object.

\subsection*{Data post-processing and standardisation}
\label{subsec:postprocessing}
After acquisition and annotation, all images underwent standardised preprocessing to ensure consistency and reproducibility across different \ac{us} devices and acquisition protocols. 

\subsubsection*{Annotation protocol}
Two complementary annotation approaches were used across datasets. This methodological difference (manual vs. ellipse-derived) contributes to the observed performance variations and highlights the importance of harmonising the annotation protocol in multi-centre datasets.

\noindent \textbf{Manual landmark annotation (FP and UCL):} Expert sonographers manually placed anatomical landmarks using the \ac{via} tool~\cite{dutta2019vgg}, following ISUOG guidelines~\cite{salomon2019isuog}. FP annotations were performed by senior maternal-fetal clinicians (inter-rater agreement 93--95\%~\cite{burgos2020zdataset}); UCL annotations were performed by a clinical research fellow under the supervision of a senior consultant. Each measurement requires two points marking the start and end of the biometric diameter; on average, each plane annotation for landmarks took 20 seconds, which is lower than the 70 seconds required for manual structure segmentation delineation~\cite{bano2021autofb}.

\noindent \textbf{Ellipse-derived landmarks (HC18):} Landmarks were automatically extracted from expert-annotated \ac{hc} segmentation masks via least-squares ellipse fitting~\cite{fitzgibbon1996ellipsefit}. \ac{bpd} and \ac{ofd} measurements were derived from the ellipse's minor and major axes, respectively. Inter-observer variability in the HC18 test set (experienced sonographer vs. medical researcher) showed \ac{ga} estimation differences of 0.8--1.6 days, highlighting the inherent variability even between trained observers~\cite{van2018automated}. This ellipse-fitting approach differs from clinical caliper placement conventions, introducing potential systematic bias in HC18 annotations.

All landmark coordinates are provided in pixel coordinates, with the origin (0, 0) at the top-left corner of the image. Coordinates are provided as floating-point values to maintain sub-pixel precision after scaling operations. Quality control measures ensured annotation accuracy: i) plane verification: each image verified to show the correct \ac{sp} for the intended measurement; ii) anatomical checks: presence of required landmarks confirmed (e.g., cavum septum pellucidum for head, stomach bubble for abdomen); iii) measurement validation: computed measurements checked against expected ranges for gestational age; iv) outlier detection: automated checks identified potential annotation errors for manual review.

\subsubsection*{Preprocessing and training setup}

All \ac{us} images underwent standard preprocessing before model training and evaluation to ensure consistency across datasets and compatibility with downstream \ac{dl} frameworks. Each image was first converted to grayscale and cropped to remove any textual overlays, scale bars, or machine-interface elements, following the same strategy used in AutoFB~\cite{bano2021autofb}. Pixel intensities were normalised to the range [0,1] by linear min–max scaling. 

For model input, the HRNet-W18 landmark detection model requires fixed-size 256~$\times$~256 pixel inputs. For each image, we compute preprocessing parameters as follows: the crop centre is set to the image centre ($w/2$, $h/2$), and the scale factor is computed as $\texttt{scale} = \max(w, h) / (1.7 \times 256)$, where $w$ and $h$ are the image dimensions. This ensures the entire image content is visible within the crop while maintaining consistent preprocessing across datasets. The HC18 dataset was originally preprocessed using ellipse-centric parameters (centre at the ellipse centroid, scale derived from the ellipse dimensions), which resulted in 84.5\% of landmarks falling outside the 64$\times$64 heatmap space in which targets are generated. We recomputed HC18 preprocessing using the image-centric formula above, reducing out-of-bounds landmarks to 0\%. The same procedure was applied uniformly across all datasets, and occasional out-of-bounds cases in FP and UCL were automatically corrected by recomputing the crop parameters.

Cross-dataset training revealed that different datasets employed opposite endpoint conventions for specific measurements. To ensure consistent supervision, we leveraged BiometryNet's \ac{dod} mechanism, which learns a measurement-specific orientation vector $\mathbf{d\_vect}$ during training and extended the framework to save $\mathbf{d\_vect}$ and apply it at inference time, which automatically corrects predicted landmarks to match the training dataset's endpoint ordering convention. This approach avoids manual point-swapping and instead lets the model learn and remember the appropriate orientation for each measurement. Without this correction, cross-dataset errors were artificially inflated because many predictions were swapped endpoint pairs that represented correct localisation but mismatched ordering conventions.

For data augmentation, random transformations were applied during training to increase robustness to operator- and device-related variations. These included random rotations within $\pm20^{\circ}$, horizontal and vertical flips, random scaling (90--110\%), translation (up to 10\% of the image size), and brightness/contrast jitter (up to 20\%). All augmentation parameters and intensity normalisation follow the settings reported in AutoFB~\cite{bano2021autofb}, which demonstrated effective regularisation for heterogeneous \ac{us} data. 

\subsection*{Scale recovery methodology}
\label{subsec:scale}

Conversion from pixels to millimetres is required to obtain accurate measurements that can be compared with those obtained in the clinic. While this information is usually available during an examination or is embedded in the raw image data, some retrospectively collected images may lack it. Therefore, we perform scale recovery using the approach presented in AutoFB~\cite{bano2021autofb}. It recovers the true scale by detecting ruler markers using template matching. For Aloka devices, where not all ruler markers are displayed, we use a different approach: detecting markers on the lower vertical band, filtering their median frequency, and estimating the scale from the median inter-tick pixel spacing. For this dataset, scale recovery was successfully performed for all images (\texttt{px\_to\_mm\_rate} field of the annotation files). Images were resized to the acquisition device's standard dimensions to maintain consistent spatial resolution during preprocessing.

\bibliography{sample}

@inproceedings{bano2021autofb,
  title={AutoFB: Automating Fetal Biometry Estimation from Standard Ultrasound Planes},
  author={Bano, Sophia and Dromey, Brian and Vasconcelos, Francisco and Napolitano, Raffaele and David, Anna L and Peebles, Donald M and Stoyanov, Danail},
  booktitle={Proc Int Conf on Medical Image Computing and Computer Assisted Intervention},
  pages={228--238},
  year={2021},
  organization={Springer}
}

@article{dromey2020dimensionless,
  title={Dimensionless squared jerk: An objective differential to assess experienced and novice probe movement in obstetric ultrasound},
  author={Dromey, Brian P and Ahmed, Shahanaz and Vasconcelos, Francisco and Mazomenos, Evangelos and Kunpalin, Yada and Ourselin, Sebastien and Deprest, Jan and David, Anna L and Stoyanov, Danail and Peebles, Donald M},
  journal={Prenatal Diagnosis},
  year={2020},
  publisher={Wiley Online Library}
}

@article{salomon2008score,
  title={A score-based method for quality control of fetal images at routine second-trimester ultrasound examination},
  author={Salomon, LJ and Winer, N and Bernard, JP and Ville, Y},
  journal={Prenatal Diagnosis},
  volume={28},
  number={9},
  pages={822--827},
  year={2008},
  publisher={Wiley Online Library}
}

@article{joskowicz2019observervar,
  title={Inter-observer variability of manual contour delineation of structures in CT},
  author={Joskowicz, Leo and Cohen, D and Caplan, N and Sosna, J},
  journal={European radiology},
  volume={29},
  number={3},
  pages={1391--1399},
  year={2019},
  publisher={Springer}
}

@article{burgos2020zdataset,
  title={Evaluation of deep convolutional neural networks for automatic classification of common maternal fetal ultrasound planes},
  author={Burgos-Artizzu, Xavier P and Coronado-Guti{\'e}rrez, David and Valenzuela-Alcaraz, Brenda and Bonet-Carne, Elisenda and Eixarch, Elisenda and Crispi, Fatima and Gratac{\'o}s, Eduard},
  journal={Sci. Rep.},
  volume={10},
  number={1},
  pages={1--12},
  year={2020},
  publisher={Nature Publishing Group}
}

@inproceedings{avisdris2022biometrynet,
  title={BiometryNet: Landmark-based Fetal Biometry Estimation from Standard Ultrasound Planes},
  author={Avisdris, Netanell and Joskowicz, Leo and Dromey, Brian and David, Anna L and Peebles, Donald M and Stoyanov, Danail and Ben Bashat, Dafna and Bano, Sophia},
  booktitle={International Conference on Medical Image Computing and Computer-Assisted Intervention},
  pages={279--289},
  year={2022},
  organization={Springer}
}

@article{salomon2019isuog,
  title={ISUOG Practice Guidelines: ultrasound assessment of fetal biometry and growth},
  author={Salomon, LJ and Alfirevic, Z and Da Silva Costa, F and Deter, RL and Figueras, F and Ghi, T et al.},
  journal={Ultrasound Obstetr Gynecol},
  volume={53},
  number={6},
  pages={715--723},
  year={2019},
  publisher={Wiley Online Library}
}

@article{khan2017bpdfl,
  title={Automatic Detection and Measurement of Fetal Biparietal Diameter and Femur Length—Feasibility on a Portable Ultrasound Device},
  author={Khan, Naiad Hossain and Tegnander, Eva and Dreier, Johan Morten and Eik-Nes, Sturla and Torp, Hans and Kiss, Gabriel},
  journal={Open J Obstetr Gynecol},
  volume={7},
  number={3},
  pages={334--350},
  year={2017},
}

@article{van2018automated,
  title={Automated measurement of fetal head circumference using 2D ultrasound images},
  author={van den Heuvel, Thomas LA and de Bruijn, Dagmar and de Korte, Chris L and Ginneken, Bram van},
  journal={PloS One},
  volume={13},
  number={8},
  pages={e0200412},
  year={2018},
}

@inproceedings{al2019bpdauto,
  title={Improving fetal head contour detection by object localisation with deep learning},
  author={Al-Bander, Baidaa and Alzahrani, Theiab and Alzahrani, Saeed and Williams, Bryan M and Zheng, Yalin},
  booktitle={Annual Conference on Medical Image Understanding and Analysis},
  pages={142--150},
  year={2019},
  organization={Springer}
}

@article{sarris2012intra,
  title={Intra-and interobserver variability in fetal ultrasound measurements},
  author={Sarris, I and Ioannou, C and Chamberlain, P and Ohuma, E and Roseman, F and Hoch, L and Altman, DG and Papageorghiou, AT and International Fetal and Newborn Growth Consortium for the 21st Century (INTERGROWTH-21st)},
  journal={Ultrasound Obstetr Gynecol},
  volume={39},
  number={3},
  pages={266--273},
  year={2012},
  publisher={Wiley Online Library}
}

@article{zhang2017automatic_fhc,
  title={Automatic image quality assessment and measurement of fetal head in two-dimensional ultrasound image},
  author={Zhang, Lei and Dudley, Nicholas J and Lambrou, Tryphon and Allinson, Nigel and Ye, Xujiong},
  journal={J Med Imaging (Bellingham)},
  volume={4},
  number={2},
  pages={024001},
  year={2017},
  publisher={International Society for Optics and Photonics}
}

@inproceedings{plotka2021fetalnet,
  title={FetalNet: Multi-task deep learning framework for fetal ultrasound biometric measurements},
  author={P{\l}otka, Szymon and W{\l}odarczyk, Tomasz and Klasa, Adam and Lipa, Micha{\l} and Sitek, Arkadiusz and Trzci{\'n}ski, Tomasz},
  booktitle={International Conference on Neural Information Processing},
  pages={257--265},
  year={2021},
  organization={Springer}
}

@inproceedings{prieto2021automated,
  title={An automated framework for image classification and segmentation of fetal ultrasound images for gestational age estimation},
  author={Prieto, Juan C and Shah, Hina and Rosenbaum, Alan J and Jiang, Xiaoning and Musonda, Patrick and Price, Joan T and Stringer, Elizabeth M and Vwalika, Bellington and Stamilio, David M and Stringer, Jeffrey SA},
  booktitle={Medical Imaging 2021: Image Processing},
  volume={11596},
  pages={115961N},
  year={2021},
  organization={International Society for Optics and Photonics}
}

@article{torres2022headus_review,
  title={A review of image processing methods for fetal head and brain analysis in ultrasound images},
  author={Torres, Helena R and Morais, Pedro and Oliveira, Bruno and Birdir, Cahit and R{\"u}diger, Mario and Fonseca, Jaime C and Vila{\c{c}}a, Jo{\~a}o L},
  journal={Comput. Methods Programs Biomed.},
  pages={106629},
  year={2022},
  publisher={Elsevier}
}

@article{Fiorentino2022fetalreview,
title = {A review on deep-learning algorithms for fetal ultrasound-image analysis},
journal = {Medical Image Analysis},
volume = {83},
pages = {102629},
year = {2023},
issn = {1361-8415},
doi = {https://doi.org/10.1016/j.media.2022.102629},
url = {https://www.sciencedirect.com/science/article/pii/S1361841522002572},
author = {Maria Chiara Fiorentino and Francesca Pia Villani and Mariachiara {Di Cosmo} and Emanuele Frontoni and Sara Moccia},
}

@article{menze2014multimodal,
  title={The multimodal brain tumor image segmentation benchmark (BRATS)},
  author={Menze, Bjoern H and Jakab, Andras and Bauer, Stefan and Kalpathy-Cramer, Jayashree and Farahani, Keyvan and Kirby, Justin and Burren, Yuliya and Porz, Nicole and Slotboom, Johannes and Wiest, Roland and others},
  journal={IEEE transactions on medical imaging},
  volume={34},
  number={10},
  pages={1993--2024},
  year={2014},
  publisher={IEEE}
}

@article{payette2021automatic,
  title={An automatic multi-tissue human fetal brain segmentation benchmark using the fetal tissue annotation dataset},
  author={Payette, Kelly and de Dumast, Priscille and Kebiri, Hamza and Ezhov, Ivan and Paetzold, Johannes C and Shit, Suprosanna and Iqbal, Asim and Khan, Romesa and Kottke, Raimund and Grehten, Patrice and others},
  journal={Scientific Data},
  volume={8},
  number={1},
  pages={1--14},
  year={2021},
  publisher={Nature Publishing Group}
}

@inproceedings{fitzgibbon1996ellipsefit,
  title={Direct least squares fitting of ellipses},
  author={Fitzgibbon, Andrew W and Pilu, Maurizio and Fisher, Robert B},
  booktitle={Proc Int Conf on Pattern Recognition},
  volume={1},
  pages={253--257},
  year={1996},
  organization={IEEE}
}

@article{zhu2021automatic,
  title={Automatic measurement of fetal femur length in ultrasound images: a comparison of random forest regression model and SegNet},
  author={Zhu, Fengcheng and Liu, Mengyuan and Wang, Feifei and Qiu, Di and Li, Ruiman and Dai, Chenyang},
  journal={Mathematical Biosciences and Engineering},
  volume={18},
  number={6},
  pages={7790--7805},
  year={2021},
  publisher={AIMS Press}
}

@inproceedings{sinclair2018human,
  title={Human-level performance on automatic head biometrics in fetal ultrasound using fully convolutional neural networks},
  author={Sinclair, Matthew and Baumgartner, Christian F and Matthew, Jacqueline and Bai, Wenjia and Martinez, Juan Cerrolaza and Li, Yuanwei and Smith, Sandra and Knight, Caroline L and Kainz, Bernhard and Hajnal, Jo and others},
  booktitle={2018 40th annual international conference of the IEEE engineering in medicine and biology society (EMBC)},
  pages={714--717},
  year={2018},
  organization={IEEE}
}

@article{oghli2021automatic,
  title={Automatic fetal biometry prediction using a novel deep convolutional network architecture},
  author={Oghli, Mostafa Ghelich and Shabanzadeh, Ali and Moradi, Shakiba and Sirjani, Nasim and Gerami, Reza and Ghaderi, Payam and Taheri, Morteza Sanei and Shiri, Isaac and Arabi, Hossein and Zaidi, Habib},
  journal={Physica Medica},
  volume={88},
  pages={127--137},
  year={2021},
  publisher={Elsevier}
}

@inproceedings{dutta2019vgg,
  author = {Dutta, Abhishek and Zisserman, Andrew},
  title = {The {VIA} Annotation Software for Images, Audio and Video},
  booktitle = {Proceedings of the 27th ACM International Conference on Multimedia},
  series = {MM '19},
  year = {2019},
  isbn = {978-1-4503-6889-6/19/10},
  location = {Nice, France},
  numpages = {4},
  url = {https://doi.org/10.1145/3343031.3350535},
  doi = {10.1145/3343031.3350535},
  publisher = {ACM},
  address = {New York, NY, USA},
}

@article{kim2019automatic,
  title={Automatic evaluation of fetal head biometry from ultrasound images using machine learning},
  author={Kim, Hwa Pyung and Lee, Sung Min and Kwon, Ja-Young and Park, Yejin and Kim, Kang Cheol and Seo, Jin Keun},
  journal={Physiological measurement},
  volume={40},
  number={6},
  pages={065009},
  year={2019},
  publisher={IOP Publishing}
}

@inproceedings{oghli2020automatic,
  title={Automatic measurement of fetal head biometry from ultrasound images using deep neural networks},
  author={Oghli, Mostafa Ghelich and Moradi, Shakiba and Sirjani, Nasim and Gerami, Reza and Ghaderi, Payam and Shabanzadeh, Ali and Arabi, Hossein and Shiri, Isaac and Zaidi, Habib},
  booktitle={2020 IEEE Nuclear Science Symposium and Medical Imaging Conference (NSS/MIC)},
  pages={1--3},
  year={2020},
  organization={IEEE}
}

@article{slimani2023fetal,
  title={Fetal biometry and amniotic fluid volume assessment end-to-end automation using Deep Learning},
  author={Slimani, Saad and Hounka, Salaheddine and Mahmoudi, Abdelhak and Rehah, Taha and Laoudiyi, Dalal and Saadi, Hanane and Bouziyane, Amal and Lamrissi, Amine and Jalal, Mohamed and Bouhya, Said and others},
  journal={Nature Communications},
  volume={14},
  number={1},
  pages={7047},
  year={2023},
  publisher={Nature Publishing Group UK London}
}

@article{lee2023machine,
  title={Machine learning for accurate estimation of fetal gestational age based on ultrasound images},
  author={Lee, Lok Hin and Bradburn, Elizabeth and Craik, Rachel and Yaqub, Mohammad and Norris, Shane A and Ismail, Leila Cheikh and Ohuma, Eric O and Barros, Fernando C and Lambert, Ann and Carvalho, Maria and others},
  journal={NPJ digital medicine},
  volume={6},
  number={1},
  pages={36},
  year={2023},
  publisher={Nature Publishing Group UK London}
}

@article{venturini2025whole,
  title={Whole examination AI estimation of fetal biometrics from 20-week ultrasound scans},
  author={Venturini, Lorenzo and Budd, Samuel and Farruggia, Alfonso and Wright, Robert and Matthew, Jacqueline and Day, Thomas G and Kainz, Bernhard and Razavi, Reza and Hajnal, Jo V},
  journal={NPJ Digital Medicine},
  volume={8},
  number={1},
  pages={22},
  year={2025},
  publisher={Nature Publishing Group UK London}
}

@article{benson2025fetal,
  title={Fetal gestational age estimation using artificial intelligence on non-targeted ultrasound images and video},
  author={Benson, Martin and Walton, Sacha and Hartley, Tom and Meagher, Simon and Seshadri, Suresh and Sleep, Nicholas and Papageorghiou, Aris T},
  journal={npj Digital Medicine},
  volume={8},
  number={1},
  pages={700},
  year={2025},
  publisher={Nature Publishing Group UK London}
}

@article{goetz2025development,
  title={Development and temporal validation of a deep learning model for automatic fetal biometry from ultrasound videos},
  author={Goetz-Fu, M and Haller, M and Collins, T and Begusic, N and Jochum, F and Keeza, Y and Uwineza, J and Marescaux, J and Weingertner, AS and Sanan{\`e}s, N and others},
  journal={Journal of Gynecology Obstetrics and Human Reproduction},
  pages={103039},
  year={2025},
  publisher={Elsevier}
}

\section*{Acknowledgements}
This work was supported in whole, or in part, by the Wellcome/EPSRC Centre for Interventional and Surgical Sciences [203145/Z/16/Z], the Engineering and Physical Sciences Research Council (EPSRC) [EP/P027938/1, EP/R004080/1, EP/P012841/1, NS/A000027/1], Wellcome [WT101957], the Royal Academy of Engineering under the Chair in Emerging Technologies programme (CiET1819/2/36), the Horizon 2020 FET Open [863146], and the Kamin Grants [63418, 72126] from the Israel Innovation Authority. We acknowledge the support of D. Casagrandi and P. Pandya in acquiring the data used in this study.

\section*{Author contributions statement}
C.D.V.\ performed experiments, analysed data and results, and wrote the manuscript. Z.M.\ contributed to the analysis of the results and manuscript writing. N.A.\ developed the BiometryNet framework used for technical validation. B.D. and R.N.\ contributed to UCL data collection. D.B.B., F.V., D.S., and L.J.\ provided supervision and edited the manuscript. S.B.\ conceived the study, supervised the project, and edited the manuscript. All authors reviewed and approved the final manuscript.

\section*{Additional information}
\textbf{Accession codes:} All data and annotations are available on the \href{https://doi.org/10.5522/04/30819911d}{UCL Research Data Repository}. Training code and evaluation pipelines are available at \href{https://github.com/surgical-vision/Multicentre-Fetal-Biometry.git}{https://github.com/surgical-vision/Multicentre-Fetal-Biometry.git}.\\
\noindent \textbf{Competing interests:} The authors declare no competing interests.

\end{document}